\documentclass[10pt,twocolumn,letterpaper]{article}

\usepackage{iccv}
\usepackage{times}
\usepackage{epsfig}
\usepackage{graphicx}
\usepackage{amsmath}
\usepackage{amssymb}

\usepackage{booktabs}
\usepackage{bbm}
\usepackage{xcolor}
\usepackage{multicol}

\usepackage[format=plain,labelformat=simple,labelsep=period,font=small,compatibility=false]{caption}
\usepackage[font=footnotesize,skip=3pt,subrefformat=parens]{subcaption}

\usepackage[pagebackref=true,breaklinks=true,letterpaper=true,colorlinks,bookmarks=false]{hyperref}

\usepackage[capitalize]{cleveref}

\iccvfinalcopy % *** Uncomment this line for the final submission

\def\iccvPaperID{7} % *** Enter the ICCV Paper ID here

\ificcvfinal\fi

\begin{document}

%%%%%%%%% TITLE
\title{Consistency Regularization for Generalizable Source-free Domain Adaptation}

\author{Longxiang Tang$^1$\qquad
Kai Li$^2$\qquad
Chunming He$^1$\qquad
Yulun Zhang$^3$\qquad
Xiu Li$^1$ \\
$^1$Tsinghua Shenzhen International Graduate School, Tsinghua University \\
$^2$NEC Laboratories America \qquad
$^3$ETH Zurich, Switzerland \\
{\tt\small \{lloong.x, li.gml.kai, chunminghe19990224, yulun100\}@gmail.com} \\
{\tt\small li.xiu@sz.tsinghua.edu.cn}
% \and
}

\maketitle
\ificcvfinal\thispagestyle{empty}\fi

%%%%%%%%% ABSTRACT
\begin{abstract}
    % % before chatGPT
    % Source-free domain adaptation (SFDA) aims to adapt a well-trained source model to an unlabelled target domain without accessing source dataset, which is practical in various real-world scenarios. Existing SFDA methods ONLY evaluate their adapted model on the target training set while ignore the data from unseen but identically distributed testing set, thus leading to a overfitting problem and restricting the generalization ability. In this paper, we propose a consistency regularization framework to realize a generalizable SFDA and significantly boost model performance on both target training and testing datasets. Specifically, we generate soft pseudo-labels from the weak-augmented images and supervise the strong-augmented ones, avoiding overfitting issues and enhance model generalization abilities. To leverage more potentially useful supervision, we present a sampling-based pseudo-label selection strategy, take samples with severer domain shift into consideration. Moreover, global-oriented calibration methods are introduced to exploit global class distribution and feature clusters information, thereby further improving adaptation process. Extensive experiments demonstrate our method achieves state-of-the-art performance on several DA benchmarks, especially on those unseen testing data.
    % after chatGPT
    Source-free domain adaptation (SFDA) aims to adapt a well-trained source model to an unlabelled target domain without accessing the source dataset, making it applicable in a variety of real-world scenarios. Existing SFDA methods ONLY assess their adapted models on the target training set, neglecting the data from unseen but identically distributed testing sets. This oversight leads to overfitting issues and constrains the model's generalization ability. In this paper, we propose a consistency regularization framework to develop a more generalizable SFDA method, which simultaneously boosts model performance on both target training and testing datasets. Our method leverages soft pseudo-labels generated from weakly augmented images to supervise strongly augmented images, facilitating the model training process and enhancing the generalization ability of the adapted model. To leverage more potentially useful supervision, we present a sampling-based pseudo-label selection strategy, taking samples with severer domain shift into consideration. Moreover, global-oriented calibration methods are introduced to exploit global class distribution and feature cluster information, further improving the adaptation process. Extensive experiments demonstrate our method achieves state-of-the-art performance on several SFDA benchmarks, and exhibits robustness on unseen testing datasets.
\end{abstract}

%%%%%%%%% BODY TEXT
\section{Introduction}
\label{sec:intro}

Deep neural networks achieve impressive success on various tasks~\cite{he2016deep, dosovitskiy2020image, devlin2018bert,li2020adversarial} while suffering from performance degradation when applied to data with a different distribution, which is called domain shift~\cite{wang2018deep}. 
\textit{Unsupervised Domain Adaptation} (UDA) approaches provide a promising solution to address this issue by learning an adaptive model jointly with labeled images (source domain) and unlabeled images with shifted distribution (target domain). Most existing approaches~\cite{rozantsev2018beyond,ganin2016domain,tang2020unsupervised,xu2021cdtrans} focus on inter-domain feature alignment by leveraging both domain data distribution, which implies an availability to source domain data when performing an adaptation strategy. 
However, considering real-world scenarios, source datasets storage costs and privacy issues may limit the accessibility to source data. Source datasets for pretraining are always on a large scale, which makes it infeasible to deploy traditional UDA methods to portable devices like mobile phones, though source model size is reasonable~\cite{liang2020we}. Meanwhile, source data might be privacy sensitive and related to copyright restrictions, like medical images and commercial data.

\begin{figure}[t]
	\centering
	\setlength{\abovecaptionskip}{0cm}
	\begin{subfigure}[b]{0.49\linewidth}
		\centering
		\includegraphics[width=\linewidth]{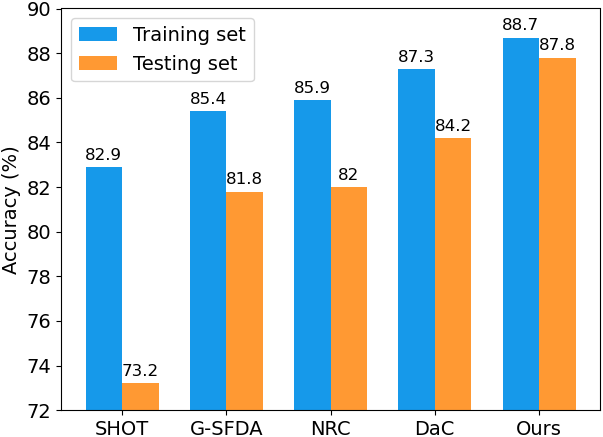}
		\label{fig1-1}
	\end{subfigure}
	\hfill
	\begin{subfigure}[b]{0.49\linewidth}
		\centering
		\includegraphics[width=\linewidth]{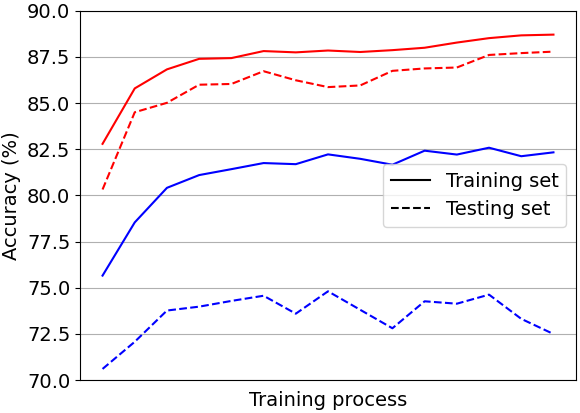}
		\label{fig1-2}
	\end{subfigure}
	\begin{subfigure}[b]{\linewidth}
		\centering
		\includegraphics[width=\linewidth]{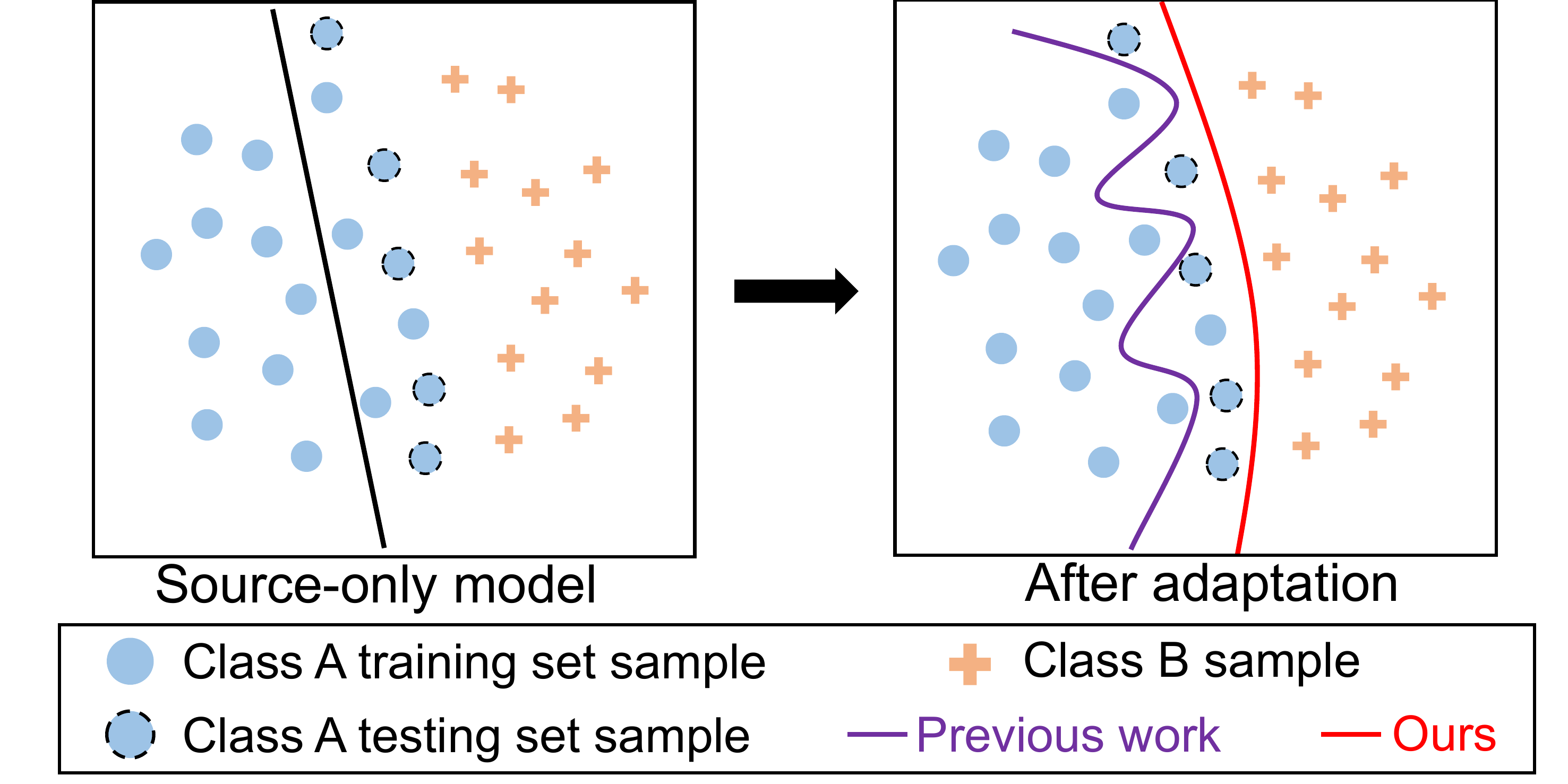}
		\label{fig1-3}
	\end{subfigure}
 \vspace{-5mm}
\caption{(\textbf{Top left}) Accuracy (\%) comparison with existing methods~\cite{liang2020we,yang2021exploiting,yang2021generalized} on target training and testing data of VisDA-2017 dataset~\cite{peng2017visda}. (\textbf{Top right}) Training curve. Our results are in {\color{red} red}, while SHOT~\cite{liang2020we} is in {\color{blue} blue}. (\textbf{Bottom}) Diagram of adaptation effect comparison. Existing SFDA methods only evaluate and improve their performance on target training data, which shows a performance degradation on unseen but identically distributed data. Our approaches construct a generalizable model, boosting the performance both on training and testing data.}
	\label{fig1}
	\vspace{-4mm}
\end{figure}

To this end, recently a new setting of UDA which alleviates the need to access source data, called \textit{Source-free Domain Adaptation} (SFDA), has drawn increasing research interest.
SHOT~\cite{liang2020we} performs a source hypothesis transferring by fixing the classifier and adopting information maximization loss, NRC~\cite{yang2021exploiting} and G-SFDA~\cite{yang2021generalized} encourage similar prediction among nearest neighbors, and SFDA-DE~\cite{ding2022source} aligns surrogate features from estimated source distribution to target features. 
These existing paradigms have shown promising results under a transductive learning manner, where the models are trained and tested on the same dataset, which often leads to overfitting problems~\cite{huang2021rda}. 

However, considering the real-world application, it's impractical to collect all target data for model training. So the primary object of SFDA is to perform better on unseen target domain data, instead of only being effective on the training set. Thus, it is essential to propose an overfitting-robust model with enhanced generalization ability.
% original: Though achieved some success, none of existing paradigms considered the manifold smoothness assumption~\cite{cayton2005algorithms} and might overfit to target data as performing a transductive learning manner.
% SHOT等没考虑smooth, NRC等neighbor可能noisy, minority类的neighbor可能sparse, 我们的生成的data point更靠近，且模拟unseen的样本

Shown in the top row of \cref{fig1}, models trained with previous works are suffering from performance degradation when handling unseen but identically distributed images, i.e., the testing data. As illustrated in the bottom part, we argue that the reason for previous works who perform well on training set but fail to classify testing samples, is that they only consider existing target samples and construct an unsmooth model manifold~\cite{cayton2005algorithms}.
In this case, overfitting issues easily occur during the transductive learning process, especially since there's no external supervision signal under the SFDA setting. Although neighborhood-based methods~\cite{yang2021exploiting,yang2021generalized} show some insights to constrain the local consistency by nearest neighbors, model manifold smoothness still cannot be guaranteed due to potentially spare feature space and limited target samples.

To address the above issues, in this paper, we propose a consistency regularization framework to realize a generalizable SFDA. 
\underline{(1)} To avoid overfitting and promote the model's generalization ability, we enforce consistent predictions between target data and its perturbation form derived by image augmentation. It's implemented by generating soft pseudo-labels with model outputs of weakly augmented images, and supervising the strongly augmented ones. With this, we could construct a smooth model manifold which is beneficial to learn a transferable model and avoid overfitting issues.
\underline{(2)} We argue that pseudo-label selection is crucial for avoiding error accumulation because no explicit supervision is provided under the SFDA setting. Different from previous thresholding methods who may omit valuable intrinsic information with a rough threshold, we present a sampling-based pseudo-label selection method, filtering samples with probability associated with their prediction confidence metric, which leverages potentially useful samples.
\underline{(3)} We also introduce global information to calibrate the point-wise operations. We leverage class prediction distribution to apply class-wise weighting, preventing class imbalance issues without introducing any separate class balance loss. To identify the noisy outliers, feature cluster information is integrated by calculating class prototypes and selecting reliable training samples.

% ver.2
% Besides, global information is also introduced to calibrate the point-wise operations. To prevent class imbalance issues, we apply class-wise weights based on current learning status without introducing any separate class balance loss, which is more robust and effective towards data with intrinsic class imbalance. To identify the outliers, global feature cluster relations are considered by calculating class prototypes and selecting reliable training samples.

% ver.1
% Besides, to prevent class imbalance issues, we apply class-wise weights based on current learning status without introducing any separate class balance loss. It's shown that our solution is more robust and effective towards data with intrinsic class imbalance. 
% Consistency regularization only explores local information, thus we also include global relations to advance model performance by calculating class prototypes and identifying the outliers.
% constrain model predictions to be invariant to input noise

Our contributions can be summarized as follows:
%创新点都缩到五行以下
\begin{itemize}
 \vspace{-3pt}
	\item We propose the consistency regularization framework under the SFDA setting for the first time, generating a robust and generalizable model. It effectively avoids overfitting issues and boosts performance both on target training and testing sets.
	% We propose a consistency regularization framework to realize a generalizable SFDA. To the best of our knowledge, it is the first attempt to explore the potential of consistency regularization in the SFDA task. 
    % By observing and analyzing the model inconsistency and lack of generalization issues under a source-free domain adaptation setting, we first introduce the consistency regularization method to boost the adaptation performance, avoid overfitting and improve the model robustness towards unseen data.
 \vspace{-3pt}
	\item To advance the effect of consistency regularization, we propose the sampling-based pseudo-label selection strategy to leverage more potential samples and the global-oriented calibration techniques to constrain and enhance the adaptation process.
	% We propose a sampling-based method to effectively leverage more potential samples and extract latent information under the SFDA setting. Besides, global-oriented calibration techniques are presented to enhance the consistency regularization training process.
	% 	\item We propose a sampling-based method to select useful samples for training, which can effectively leverage more potential samples and extract latent information under the SFDA setting. Besides, class-wise weighting and prototype techniques are presented as global-oriented calibration components to enhance the consistency regularization process.
 \vspace{-3pt}
	\item Massive experiments show that our proposed method outperforms existing approaches on DA benchmarks, especially with a large margin on target testing set.
\end{itemize}

\section{Related Work}
\label{sec:related}

\noindent \textbf{Unsupervised Domain Adaptation.} 
UDA attempts to learn an adaptive model jointly with labeled images (source domain) and unlabeled images (target domain)~\cite{wang2018deep}, which can erase the burden of labeling target domain data. 
% Being researched for many years,
These years
UDA has achieved great success on different tasks~\cite{xu2021cdtrans,deng2021unbiased,zou2018unsupervised}. 
Most common existing UDA approaches are trying to align feature distribution between source and target domain, for instance, Maximum Mean Discrepancy (MMD)~\cite{gretton2006kernel} based methods~\cite{rozantsev2018beyond,long2015learning} align feature embedding in reproducing kernel Hilbert space, which can reduce the distance of representations from each domain. Holding the same point, DANN~\cite{ganin2016domain}
% and WDGRL~\cite{shen2018wasserstein} 
achieve this alignment through an adversarial training approach, by fooling a newly added domain classifier for extracting domain-invariant feature representation. 
% In addition, AdaBN~\cite{li2018adaptive} and AutoDIAL~\cite{maria2017autodial} assume that domain knowledge is stored in batch norm statistics, and tackle the domain shift issue by transferring these statistics. 
More recently,  SRDC~\cite{tang2020unsupervised} proposes to directly uncover the intrinsic target discrimination via discriminative clustering of target data, 
% CDTrans~\cite{xu2021cdtrans} introduces a cross-attention module from famous Transformer~\cite{dosovitskiy2020image} to perform better feature alignment across noisy input pairs, 
FixBi~\cite{na2021fixbi} performs bidirectional matching and self-penalization on intermediate domains generated by fixed ratio-based mixup.
The success of these paradigms depends on the assumption of simultaneous access to source and target data, which is hard to guarantee in real-world scenarios.

\noindent \textbf{Source-free Domain Adaptation.} 
Considering the privacy and copyright restrictions when adapting models across domains, a new setting of domain adaptation, SFDA, is proposed and eliminates the requirement to source data during the adaptation procedure~\cite{liang2020we,yang2021exploiting,ding2022source,tang2023source,li2023source}. SHOT~\cite{liang2020we} freezes the source model classifier module (hypothesis) and exploits both information maximization and self-supervised pseudo-labeling to achieve feature alignment. A$^2$Net~\cite{xia2021adaptive} develops an adversarial network to seek a target-specific classifier that advances the target hard samples' recognition. Based on the observation that target data still form clear clusters with the source model, NRC~\cite{yang2021exploiting} and G-SFDA~\cite{yang2021generalized} encourage label consistency towards nearest neighbors in feature space. 
% SFDA-DE~\cite{ding2022source} estimates the source feature distribution by target data and source classifier, then samples surrogate features from it and aligns them with target features. 
BUFR~\cite{eastwood2021source} focuses on measurement shifts between domains and applies a bottom-up source feature restoration technique to boost target feature representation. DaC~\cite{zhang2022dac} divides the target data into source-like and target-specific samples, to learn global class clustering with former and learn intrinsic local structures with later. The methods mentioned above provide some insights to achieve SFDA, however, they only consider and adapt samples in the target domain in a transductive manner, which could lead to an unsmooth manifold and raise overfitting issues.

\noindent \textbf{Consistency Regularization.} 
Consistency regularization is extensively investigated in the semi-supervised learning area, and the key point is similar inputs should share consistent predictions, which can enhance the network's feature representation. 
% Temporal Ensembling~\cite{laine2016temporal} leverages outputs from different epochs and input augmentation conditions to form a consensus prediction, 
Mean Teacher~\cite{tarvainen2017mean} averages model weights using EMA over training steps and enforce consistency between student and teacher models, UDA~\cite{xie2020unsupervised} proves the importance of data augmentation and provides a new effective way to noise unlabeled examples. 
% These methods only minimize the difference between two predictions, however, 
FixMatch~\cite{sohn2020fixmatch} proposes a new framework to achieve consistency regularization by generating pseudo-labels from weakly-augmented images and using them to supervise the outputs of strongly-augmented ones. These approaches are under SSL setting, which need labeled samples to guide the model training process and can't satisfy the SFDA scenarios.

\begin{figure*}[t]
    \centering
     \setlength{\abovecaptionskip}{2mm}
    \includegraphics[width=\linewidth]{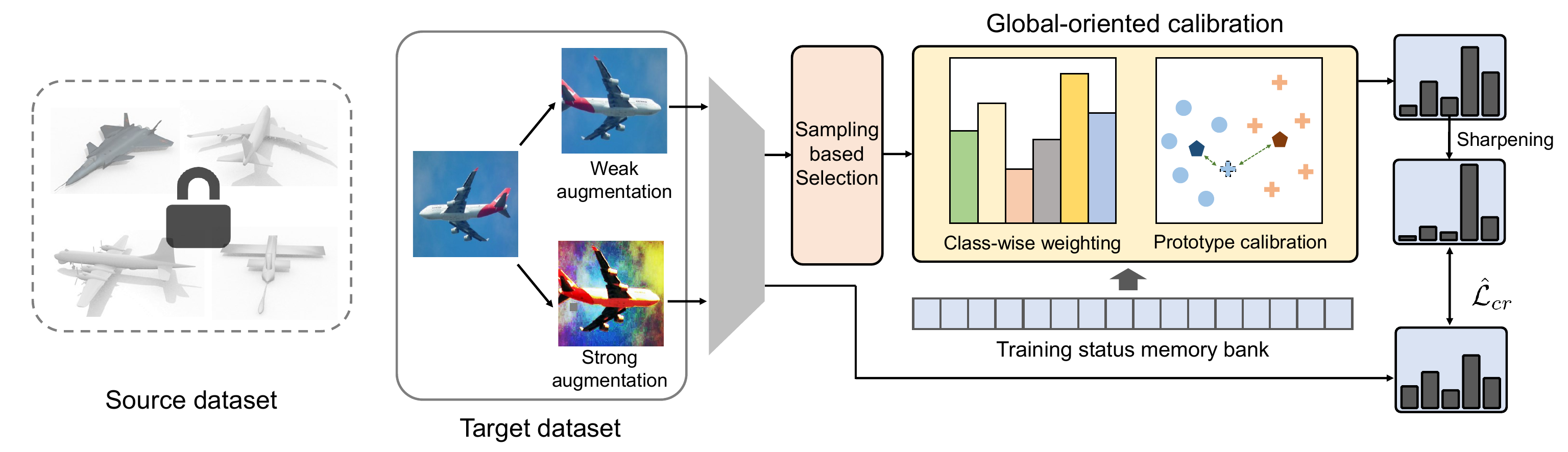} 
    \vspace{-5mm}
    \caption{Framework of our proposed SFDA method. We leverage consistency regularization to boost model generalization ability, meanwhile, sampling-based pseudo-label selection strategy and global-oriented calibration module further strengthen the adaptation process.}
    \label{fig:framework}
    \vspace{-5mm}
\end{figure*}

\section{Method}
\label{sec:method}

In this section, we formalize the definition of source-free domain adaptation and our methodology to tackle it. Firstly we introduce notations used in SFDA, then present our basic consistency regularization framework aiming to develop a more robust and generalizable model. For further boosting the performance, a sampling-based pseudo-label selection strategy is proposed and analyzed in detail. Lastly, we leverage global information to avoid degradation by performing global-oriented calibration methods.

\subsection{Preliminaries}

For the vanilla UDA classification task, we are given a source dataset $D_s={(x_i^s, y_i^s)}_{i=1}^{n_s}$ with $n_s$ labeled training samples and a target dataset $D_t={x_i^t}_{i=1}^{n_t}$ with $n_t$ unlabeled samples, where $x_s^i \in \mathcal{X}_s, x_t^i \in \mathcal{X}_t$ and $\mathcal{X}_s$ holds different underlying marginal distribution with $\mathcal{X}_t$ as domain shift exists. Under the close-set~\cite{saenko2010adapting} setting, source and target domain share the same label set $\mathcal{C} = \{1,2,\cdots, K\}$ under the K-way classification task, and $y \in \mathbb{R}^K$ is the one-hot label. SFDA considers the inaccessibility to source data problem in the adaptation phase, so only a source model $h=f\circ g$ pre-trained on $D_s$ and a target dataset $D_t$ without any label is provided when applying adaptation. Similar to existing works, the source model can be divided into two components: feature extractor $f$ who generates features $z_i=f(x_i)$ from input images and classifier $g$ who provides final prediction $p_i=p(x_i)=\delta(g(z_i))$ given $z$ where $\delta$ is the softmax function.

\subsection{Consistency Regularization}

Existing paradigms consider SFDA as merely a transductive learning process, in which they train and test models on the same set of samples, ignoring the robustness and generalization ability. However, in real-world applications, it's almost impossible to collect every target domain sample for applying adaptation, so we attempt to develop a generalizable model which can achieve high performance on unseen but identically distributed target data.
% raise a more realistic question: how to boost SFDA performance on unseen but identically distributed target data?

Manifold smoothness assumption, which is mostly considered in the semi-supervised learning (SSL) area~\cite{laine2016temporal,tarvainen2017mean,miyato2018virtual,xie2020unsupervised,sohn2020fixmatch}, describes that a model's output shouldn't change when realistic perturbations are occurred in data points~\cite{yang2021survey}, to construct a more reliable feature representation. We argue that SFDA, similar to SSL which mainly exploits potentially beneficial supervision signals from unlabeled data, should also satisfy this assumption to keep model consistency for a better understanding of target structure and robustness of unseen identically distributed data. 

We introduce consistency regularization terms to generate a smooth manifold. To realize perturbations on data points, strong augmentations $\mathcal{A}$ are applied to input images, simulating the noisy form of the original image. Model prediction of a strong augmented image is $p_i' = p(x_i') = \delta\left(h\left(\mathcal{A}\left(x_i\right)\right)\right)$. As $p_i'$ and $p_i$ are derived from the same semantic information, it's obvious that they should share consistent results, which means $\text{Dist}(p_i, p_i')$ is small. 

Many SSL methods~\cite{laine2016temporal,tarvainen2017mean,xie2020unsupervised} attempt to minimize this distance with massive implementations, as an auxiliary task to labeled sample loss. Since no label information is available, these discrepancy minimization methods are insufficient to develop a target adaptive model under the SFDA setting. We hold that with less noise and transformation, $p_i$ could provide supervision signals to $p_i'$ as it has a more accurate prediction result. Formally, we generate soft pseudo-label $\hat{p}_k(x_i)$ from model predictions:
\begin{equation}
	\hat{p}_k(x_i) = \frac{\exp(h_k(x_i)/T)}{\sum_{k=0}^K \exp(h_k(x_i)/T)}
	\label{eq1}
\end{equation}
where subscript $k$ denotes the $k^{th}$ class output, and $T<1$ is Softmax temperature. This operation can be regarded as a sharpening~\cite{berthelot2019mixmatch} to the original class probability output, which makes prediction closer to one-hot and provides more clear supervision. We argue that soft pseudo-labels could be more robust for noisy samples than one-hot labels which may enlarge wrong learning signals, especially when there's no external constraint under the SFDA setting.

As mentioned above, aligning $p_i'$ to $\hat{p_i}$ enhances the prediction consistency and provides supervision for domain adaptation. We apply cross-entropy loss here to accomplish alignment for sample $x_i$:
\begin{equation}
	\mathcal{L}_{cr}^{(i)} = H(\hat{p_i}, p_i') = -\sum_{k=1}^K \hat{p}_k(x_i)\log(p_k(\mathcal{A}(x_i)))
	\label{eq2}
\end{equation}
where $p_k(x_i)$ denotes the probability output of the $k^{th}$ class. Besides construct consistency regularization, strong augmentation could also be regarded as an expansion of target data space, which is an effective approach to avoid overfitting issues and promotes the model's generalization ability. Notice that our implementation shares some similarities with FixMatch~\cite{sohn2020fixmatch}, however, 1) FixMatch applies consistency regularization to mine unlabeled samples aiming for setting up a new model, but we are trying to adapt the pre-trained model to unlabeled data, 2) with labeled samples, Fixmatch generate vanilla one-hot pseudo-label, which is different from our approach, and 3) as discussed in \cref{sampling}, FixMatch needs labeled samples as anchors during the learning process, which is impractical for SFDA.

\subsection{Sampling-based pseudo-label selection}
\label{sampling}

One of the biggest challenges of SFDA is there's no explicit supervision signal when performing adaptation. It means error accumulation easily occur and causes degradation issue: if initially a sample was wrongly recognized as class k and continually involved in self-training, it will hardly be classified correctly thereafter, even if it's just close to class boundaries at the beginning due to domain shift. This fact implies that vanilla pseudo-labeling methods~\cite{lee2013pseudo,xie2020self,zhai2019s4l} who leverage all pseudo-labels are sub-optimal. Meanwhile, some works~\cite{sohn2020fixmatch,zhang2021flexmatch,li2021ecacl,he2023weakly} introduce threshold to select ``reliable'' pseudo-labels. However, these methods need a carefully designed threshold, and too low takes no effect while too high omits useful information. 

We provide insight into how to select pseudo-label and extract more potentially useful supervision by a sampling-based method. Consider an image input $x_i$ and its corresponding prediction vector $p_i$, if $p_i$ is close to one-hot, which means the model has high confidence to classify this sample, and the pseudo-label is more likely to be right. But at the same time, samples with relatively low confidence shouldn't be regarded wrong and abandoned, because they still carry some useful information. Only guided by ``right'' pseudo-labels, the model would tend to be biased and finally obtain sub-optimal results. We argue that conditionally adding relatively low-confidence samples to the self-training process could leverage more underlying distribution information and boost performance.

Formally, we sample pseudo-labels from the target dataset for training with a certain probability $\xi$ related to its prediction confidence:
\begin{equation}
	\xi_i = P(x_i \textit{ is selected}) = \mathcal{M}(\max_k p_k(x_i))
	\label{eq3}
\end{equation}
where $k= 1,2,\cdots,C$ and $\mathcal{M}$ denotes a mapping function with a range of $[0,1]$. 
As $\mathcal{M}$ is a monotonically non-decreasing function, we can guarantee that high-confident samples are more considered during adaptation, and besides, low-confident samples aren't ignored. With the sampling strategy, hard samples caused by domain shift could also provide feature structure information for adaptation.

Implementing a probability sampling strategy often requires a relatively long training procedure and more samples to converge. But as an approximation, we can substitute sampling results with the expectation of each sample's consistency regularization loss, accelerating model convergence. Specifically, consistency regularization loss under sampling strategy with expectation approximation is:
\begin{equation}
	\hat{\mathcal{L}}_{cr} = \mathbb{E}_{x_i\in D_t} \hat{\mathcal{L}}_{cr}^{(i)} = \frac{1}{n_t} \sum_{i=1}^{n_t} \xi_i \cdot H(\hat{p_i}, p_i')
	\label{eq4}
\end{equation}
where $\xi_i$ denotes the sampling probability in \cref{eq3}. Applying expectation instead of sampling results could helps the model perform better on small datasets because it introduces less random noise and makes the network less biased due to insufficient data points.

\subsection{Global-oriented calibration}
\label{Global-oriented-calibration}

Since consistency regularization only leverages predictions of single input, global distribution information is under-considered, which is crucial to avoid degradation issues and form a compact feature representation. 

Firstly, similar to previous works~\cite{liang2020we,yang2021exploiting,yang2021generalized}, SFDA paradigms easily suffer from degeneration problems where the model predicts all data to one or several specific classes, which greatly suppresses adaptation performance. Existing works~\cite{liang2020we,yang2021exploiting,yang2021generalized} usually introduce an additional loss term, called diversity loss, to manually constrain class balance for tackling this local minimum issue, termed as $\mathcal{L}_{div}=\sum_k \text{KL}(\overline{p}_k, \frac{1}{C})$ where $\overline{p}_k=\mathbb{E}_{x_i\in D_t}(p_k(x_i))$ is the mean probability output of class $k$. Although it shows some effect in avoiding trivial solution issues, it relies on a basic assumption that samples are uniformly distributed among different classes, which always fails on datasets \cite{peng2017visda,venkateswara2017deep,saenko2010adapting,he2023camouflaged,he2023HQG} and real-world scenarios.

We present a class-wise weighting method to alleviate model degeneration without introducing any additional loss term, and can fit the class-imbalance situation better. Classes with fewer samples (minority) may encounter severer domain shifts, or just because they lack samples intrinsically. So instead of roughly constraining them to have the same amount of samples, we weigh their supervision signals with their current learning status. Formally, We denote the number of samples classified to class $k$ as $\alpha_k$:
\begin{equation}
    \small
	\alpha_c = \sum_{i=1}^{n_t} \mathbbm{1}(\max_k p_k(x_i) > \tau) \cdot \mathbbm{1}(\arg\max_k p_k(x_i) = c)
	\label{eq5}
\end{equation}
where $\mathbbm{1}$ is the indicator function and $\tau$ is the selection threshold. $[\alpha_c]$ shows the class-wise distribution in the current training stage. With it we can derive loss weight for samples from class $c$:
\begin{equation}
	w_{div}(c) = \mathcal{T}\left(\frac{\alpha_c}{\max_c \alpha_c}\right)
	\label{eq6}
\end{equation}
where $\mathcal{T}$ is a monotonically decreasing mapping function. It's shown that minority class samples will obtain higher training weights and get promoted. Unlike previous works, our weighting method is adaptive for class imbalance: the majority of classes hold more samples but with low weight, producing a close scale of supervision signal to less but highly weighted samples. It can alleviate the negative impact of avoiding degeneration in previous works, which tends to compromise the accuracy of the majority classes.

Moreover, only considering single input is hard to recognize outliers, especially where domain shift exists. Clusters in feature space can provide distribution information inside each class and identify potentially wrongly classified samples~\cite{caron2018deep,zhang2021prototypical}. For all data instances classified to class $c$, we can compute the average feature representation of them, denoted as prototype $\eta_c$:
\begin{equation}
	\eta_c = \frac{\sum_{x_i\in D_t} {z_i \cdot\mathbbm{1}(\arg\max p_i = c)}}{\sum_{x_i\in D_t}{\mathbbm{1}(\arg\max p_i = c)}}
	\label{eq7}
\end{equation}

\begin{table*}[t]
	\centering
	\setlength{\abovecaptionskip}{2mm}
	\caption{Accuracy (\%) on VisDA-2017 (ResNet101-based methods). SF means source-free. We highlight the best and second-best results.}
	\resizebox{2.1\columnwidth}{!}{
		\begin{tabular}{c|c|cccccccccccc|c}
			\toprule
			Method & SF &  plane & bcycl & bus & car & horse & knife & mcycl & person & plant & sktbrd & train & truck & Avg. \\
			
			\midrule
			
			MCC~\cite{jin2020minimum} & $\times$ & 88.7 & 80.3 & 80.5 & 71.5 & 90.1 & 93.2 & 85.0 & 71.6 & 89.4 & 73.8 & 85.0 & 36.9 & 78.8 \\
			STAR~\cite{lu2020stochastic} & $\times$ & 95.0 & 84.0 & 84.6 & 73.0 & 91.6 & 91.8 & 85.9 & 78.4 & 94.4 & 84.7 & 87.0 & 42.2 & 82.7 \\
			RWOT~\cite{xu2020reliable} & $\times$ & 95.1 & 80.3 & 83.7 & 90.0 & 92.4 & 68.0 & 92.5 & 82.2 & 87.9 & 78.4 & 90.4 & 68.2 & 84.0 \\
			SE~\cite{french2017self} & $\times$ & 95.9 & 87.4 & 85.2 & 58.6 & 96.2 & 95.7 & 90.6 & 80.0 & 94.8 & 90.8 & 88.4 & 47.9 & 84.3 \\
			
			\midrule
			
			SHOT~\cite{liang2020we} & $\checkmark$ & 94.3 & 88.5 & 80.1 & 57.3 & 93.1 & 94.9 & 80.7 & 80.3 & 91.5 & 89.1 & 86.3 & 58.2 & 82.9 \\
			DIPE~\cite{wang2022exploring} & $\checkmark$ & 95.2 & 87.6 & 78.8 & 55.9 & 93.9 & 95.0 & 84.1 & 81.7 & 92.1 & 88.9 & 85.4 & 58.0 & 83.1 \\
			A$^2$Net~\cite{xia2021adaptive} & $\checkmark$ & 94.0 & 87.8 & 85.6 & 66.8 & 93.7 & 95.1 & 85.8 & 81.2 & 91.6 & 88.2 & 86.5 & 56.0 & 84.3 \\
			G-SFDA~\cite{yang2021generalized} & $\checkmark$ & 96.1 & 88.3 & 85.5 & 74.1 & \textbf{97.1} & 95.4 & 89.5 & 79.4 & 95.4 & 92.9 & 89.1 & 42.6 & 85.4 \\
			NRC~\cite{yang2021exploiting} & $\checkmark$ & 96.8 & 91.3 & 82.4 & 62.4 & 96.2 & 95.9 & 86.1 & 80.6 & 94.8 & 94.1 & \textbf{90.4} & 59.7 & 85.9 \\
			SFDA-DE~\cite{ding2022source} & $\checkmark$ & 95.3 & 91.2 & 77.5 & 72.1 & 95.7 & 97.8 & 85.5 & 86.1 & 95.5 & 93.0 & 86.3 & \textbf{61.6} & 86.5 \\
                DaC~\cite{zhang2022dac} & $\checkmark$ & 96.6 & 86.8 & 86.4 & \textbf{78.4} & 96.4 & 96.2 & 93.6 & 83.8 & \textbf{96.8} & \textbf{95.1} & 89.6 & 50.0 & \underline{87.3} \\
			Ours & $\checkmark$ & \textbf{97.4} & \textbf{91.8} & \textbf{87.6} & 78.1 & 96.6 & \textbf{99.3} & \textbf{90.6} & \textbf{87.2} & 95.6 & 94.6 & 88.9 & 57.3 & \textbf{88.7} \\
			
			\bottomrule
	\end{tabular}}
	\label{tab1}
	\vspace{-3mm}
\end{table*}

If an image's feature is closer to $\eta_A$ but classified as class $B$ where $A \neq B$, we can indicate that it may be a latent outlier, and it's inadvisable to generate a supervision signal from it. Because once its initial prediction is slightly wrong, the error would accumulate during the self-training process and harm the final result. Based on this, we can leverage the inconsistency of feature prototype assignment and direct prediction, further modify the point-wise consistency regularization loss weight:
\begin{equation}
	w_{proto}^{(i)} = \mathbbm{1}(\arg\min_c\text{dist}(z_i, \eta_c)=\arg\max p_i)
	\label{eq8}
\end{equation}
where $\text{dist}(z_i, \eta_c)$ is the distance between sample feature and prototype. \cref{eq8} can discard unreliable samples with disagreement of these two different semantics. With the prototype-based denoise adjustment, a more accurate set of samples is selected for the self-straining process. Note that this correction is performed in real-time, some initially abandoned samples may get rectified and be involved in adaptation, providing more supervision signals.

\section{Experiments}
\label{sec:exp}

% \subsection{Experimental setup}
% \label{subsec:setup}

\noindent\textbf{Datasets} We conduct experiments on three datasets:
\textbf{Office-Home}~\cite{venkateswara2017deep} is an object recognition benchmark for domain adaptation, which consists of 12 adaptation tasks with four domains, sharing 65 classes and a total of around 15,500 images. 
\textbf{VisDA-2017}~\cite{peng2017visda} is a synthetic-to-real object recognition dataset. It has 12 classes with two domains: the synthetic (source) domain contains 152K simulated images, and the real (target) domain consists of 55k real-world images. In addition, VisDA-2017 also provides 72k target domain test images, on which we evaluate model generalization ability.
\textbf{DomainNet}~\cite{peng2019moment} is a large-scale multi-source domain adaptation benchmark with 6 domains and 345 classes, which contains $\sim$0.6 million images. Following \cite{zhang2022dac,saito2019semi}, we perform SFDA with selected four domains and 126 classes, and also evaluate testing set accuracy with their data split, shown in supplementary material.

% TODO 确认一下要不要加这句
Since consistency regularization with strong image augmentation adds noises to samples, it can be regarded as a more challenging learning setting and require more data to converge, so sample-insufficient datasets like Office-31~\cite{saenko2010adapting} are not chosen for evaluation.

\noindent \textbf{Implementation details.} 
To achieve fair comparisons with existing methods, we adopt ResNet-50~\cite{he2016deep} for Office-Home, ResNet-101 for VisDA-2017, and ResNet-34 for DomainNet as the backbone. Consistent with SHOT~\cite{liang2020we} and other previous works~\cite{xia2021adaptive,yang2021exploiting}, we add two fully connected layers with weight normalization~\cite{salimans2016weight} after ResNet backbone. For model training, we adopt SGD with momentum 0.9, weight decay 1e-3, and learning rate scheduler $\gamma = \gamma_0 \cdot (1+10 \cdot p)^{-0.75}$ where $p$ is the training progress changing from 0 to 1 and $\gamma_0$ is the initial learning rate. Our source model building-up process is identical to SHOT for a fair comparison. In the adaptation phase, we set $\gamma_0=\text{4e-4}$, $\tau=0.8$ and batch size of 64 for all experiments, while our strong image augmentation is from RandAugment~\cite{cubuk2020randaugment}. 
For VisDA-2017, we set sharpen temperature $T$ to 0.5 and epoch to 30, which are 0.1 and 60 for Office-Home and DomainNet due to the different number of classes. Considering simplification, we plainly set the sampling probability mapping function as $\mathcal{M}(x) = x$ and the class-wise weighting mapping function as $\mathcal{T}(x) = 1-\log(x)$. Noting that $x > 0$ should be guaranteed for $\mathcal{T}$, we replace $\alpha_c=0$ with the smallest non-zero $\alpha_{c'}$ in \cref{eq6}. Our final adaptation loss is formulated as $\mathcal{L} = \mathbb{E}_{x_i\in D_t} \left[w_{div}w_{proto}\hat{\mathcal{L}}_{cr}\right]$.
For maintaining memory bank, we simply follow the common practice~\cite{yang2021exploiting,yang2021generalized} by saving and updating predictions and features.
% All of our experiments are run three times with different random seeds by PyTorch on one NVIDIA RTX 3090 GPU.
All experiments are run on NVIDIA 3090 GPU.

\begin{table}[t]
	\centering
	\setlength{\abovecaptionskip}{2mm}
	\caption{Accuracy (\%) and its drop between target training and testing data on VisDA-2017 dataset. $\dag$ means these results from \textit{VisDA2017 Classification Challenge}~\cite{peng2017visda} leaderboard, which are under \textbf{easier} vanilla UDA setting and may use \textbf{stronger} backbone causing \textbf{unfair} comparisons.}
	%	\resizebox{2.1\columnwidth}{!}{
		\begin{tabular}{c|c|ccc}
			\toprule
			Method & SF & Target & Test & Drop \\
			
			\midrule
			
			BUPT$^\dag$ & $\times$ & - & 85.4 & - \\
			IISC\_SML$^\dag$ & $\times$ & - & 86.4 & - \\
			NLE$^\dag$ & $\times$ &  - & \underline{87.7} & - \\
			
			\midrule
			
			SHOT~\cite{liang2020we} & $\checkmark$ & 82.9 & 73.2 & 9.7(11.7\%$\downarrow$) \\
			G-SFDA~\cite{yang2021generalized} & $\checkmark$ & 85.4 & 81.8 & 3.6(4.2\%$\downarrow$) \\
			NRC~\cite{yang2021exploiting} & $\checkmark$ & 85.9 & 82.0 & 3.9(4.5\%$\downarrow$) \\
                DaC~\cite{zhang2022dac} & $\checkmark$ & 87.3 & 84.2 & 3.1(3.6\%$\downarrow$) \\
			Ours & $\checkmark$ & \textbf{88.7} & \textbf{87.8} & {\color{red} \textbf{0.9(1.0\%$\downarrow$)}} \\
			
			\bottomrule
		\end{tabular}
		%	}
	\label{tab2}
	\vspace{-3mm}
\end{table}

\subsection{Results}

\begin{table*}[t]
	\centering
	\setlength{\abovecaptionskip}{2mm}
	% \caption{Accuracy (\%) on DomainNet dateset for ResNet34-based methods. SF means source-free. We highlight the best results and underline the second-best ones in SFDA methods.}
 	\caption{Accuracy (\%) on DomainNet (ResNet34-based methods). SF means source-free. We highlight the best and second-best results.}
	\resizebox{1.7\columnwidth}{!}{
		\begin{tabular}{c|c|ccccccc|c}
			\toprule
			Method & SF & Rw$\rightarrow$Cl & Rw$\rightarrow$Pt & Pt$\rightarrow$Cl & Cl$\rightarrow$Sk & Sk$\rightarrow$Pt & Rw$\rightarrow$Sk & Pt$\rightarrow$Rw & Avg.\\
			
			\midrule
			
			MME~\cite{saito2019semi} & $\times$ & 70.0 & 67.7 & 69.0 & 56.3 & 64.8 & 61.0 & 76.1 & 66.4 \\
			CDAN~\cite{long2018conditional} & $\times$ & 65.0 & 64.9 & 63.7 & 53.1 & 63.4 & 54.5 & 73.2 & 62.5 \\
                VDA~\cite{jin2020minimum} & $\times$ & 63.5 & 65.7 & 62.6 & 52.7 & 53.6 & 62.0 & 74.9 & 62.1 \\
                GVB~\cite{cui2020gradually} & $\times$ & 68.2 & 69.0 & 63.2 & 56.6 & 63.1 & 62.2 & 73.8 & 65.2 \\
			\midrule

                BAIT~\cite{yang2020casting} & $\checkmark$ & 64.7 & 65.4 & 62.1 & 57.1 & 61.8 & 56.7 & 73.2 & 63.0 \\
			SHOT~\cite{liang2020we} & $\checkmark$ & 67.1 & 65.1 & 67.2 & 60.4 & 63.0 & 56.3 & 76.4 & 65.1 \\
			G-SFDA~\cite{yang2021generalized} & $\checkmark$ & 63.4 & 67.5 & 62.5 & 55.3 & 60.8 & 58.3 & 75.2 & 63.3 \\
			NRC~\cite{yang2021exploiting} & $\checkmark$ & 67.5 & 68.0 & 67.8 & 57.6 & 59.3 & 58.7 & 74.3 & 64.7 \\
			DaC~\cite{zhang2022dac} & $\checkmark$ & 70.0 & 68.8 & 70.9 & 62.4 & \textbf{66.8} & 60.3 & \textbf{78.6} & \underline{68.3} \\
			Ours & $\checkmark$ & \textbf{72.2} & \textbf{69.4} & \textbf{72.1} & \textbf{63.2} & 66.7 & \textbf{63.4} & 77.3 & \textbf{69.2} \\
			
			\bottomrule
	\end{tabular}}
	\label{tab_domain_net}
	\vspace{-3mm}
\end{table*}

\begin{table*}[t]
	\centering
	\setlength{\abovecaptionskip}{2mm}
	\caption{Accuracy (\%) on Office-Home (ResNet50-based methods). SF means source-free. We highlight the best and second-best results.}
	\resizebox{2.1\columnwidth}{!}{
		\begin{tabular}{c|c|cccccccccccc|c}
			\toprule
			Method & SF &  \small{Ar$\rightarrow$Cl} & \small{Ar$\rightarrow$Pr} & \small{Ar$\rightarrow$Rw} & \small{Cl$\rightarrow$Ar} & \small{Cl$\rightarrow$Pr} & \small{Cl$\rightarrow$Rw} & \small{Pr$\rightarrow$Ar} & \small{Pr$\rightarrow$Cl} & \small{Pr$\rightarrow$Rw} & \small{Rw$\rightarrow$Ar} & \small{Rw$\rightarrow$Cl} & \small{Rw$\rightarrow$Pr} & Avg.\\
			
			\midrule
			
			GVB-GD~\cite{cui2020gradually} & $\times$ & 57.0 & 74.7 & 79.8 & 64.6 & 74.1 & 74.6 & 65.2 & 55.1 & 81.0 & 74.6 & 59.7 & 84.3 & 70.4 \\
			RSDA~\cite{gu2020spherical} & $\times$ & 53.2 & 77.7 & 81.3 & 66.4 & 74.0 & 76.5 & 67.9 & 53.0 & 82.0 & 75.8 & 57.8 & 85.4 & 70.9 \\
			TSA~\cite{li2021transferable} & $\times$ & 57.6 & 75.8 & 80.7 & 64.3 & 76.3 & 75.1 & 66.7 & 55.7 & 81.2 & 75.7 & 61.9 & 83.8 & 71.2 \\
			SRDC~\cite{tang2020unsupervised} & $\times$ & 52.3 & 76.3 & 81.0 & 69.5 & 76.2 & 78.0 & 68.7 & 53.8 & 81.7 & 76.3 & 57.1 & 85.0 & 71.3 \\
			FixBi~\cite{na2021fixbi} & $\times$ & 58.1 & 77.3 & 80.4 & 67.7 & 79.5 & 78.1 & 65.8 & 57.9 & 81.7 & 76.4 & 62.9 & 86.7 & 72.7 \\
			
			\midrule
			
			SHOT~\cite{liang2020we} & $\checkmark$ & 57.1 & 78.1 & 81.5 & 68.0 & 78.2 & 78.1 & 67.4 & 54.9 & 82.2 & 73.3 & 58.8 & 84.3 & 71.8 \\
			G-SFDA~\cite{yang2021generalized} & $\checkmark$ & 57.9 & 78.6 & 81.0 & 66.7 & 77.2 & 77.2 & 65.6 & 56.0 & 82.2 & 72.0 & 57.8 & 83.4 & 71.3 \\
			NRC~\cite{yang2021exploiting} & $\checkmark$ & 57.7 & \textbf{80.3} & 82.0 & 68.1 & \textbf{79.8} & 78.6 & 65.3 & 56.4 & 83.0 & 71.0 & 58.6 & \textbf{85.6} & 72.2 \\
			DIPE~\cite{wang2022exploring} & $\checkmark$ & 56.5 & 79.2 & 80.7 & \textbf{70.1} & \textbf{79.8} & 78.8 & 67.9 & 55.1 & \textbf{83.5} & \textbf{74.1} & 59.3 & 84.8 & 72.5 \\
			A$^2$Net~\cite{xia2021adaptive} & $\checkmark$ & 58.4 & 79.0 & 82.4 & 67.5 & 79.3 & 78.9 & \textbf{68.0} & 56.2 & 82.9 & \textbf{74.1} & 60.5 & 85.0 & \underline{72.8} \\
			SFDA-DE~\cite{ding2022source} & $\checkmark$ & \textbf{59.7} & 79.5 & 82.4 & 69.7 & 78.6 & \textbf{79.2} & 66.1 & \textbf{57.2} & 82.6 & 73.9 & 60.8 & 85.5 & \textbf{72.9} \\
                DaC~\cite{zhang2022dac} & $\checkmark$ & 59.1 & 79.5 & 81.2 & 69.3 & 78.9 & \textbf{79.2} & 67.4 & 56.4 & 82.4 & 74.0 & \textbf{61.4} & 84.4 & \underline{72.8} \\
			Ours & $\checkmark$ & 58.6 & 80.2 & \textbf{82.9} & 69.8 & 78.6 & 79.0 & 67.8 & 55.7 & 82.3 & 73.6 & 60.1 & 84.9 & \underline{72.8} \\
			
			\bottomrule
	\end{tabular}}
	\label{tab3}
	\vspace{-3mm}
\end{table*}

We evaluate our proposed method and show comparison results with existing UDA and SFDA methods. \cref{tab1,tab_domain_net,tab3} provide classification accuracy after adaptation on VisDA-2017, DomainNet and Office-Home datasets respectively. As shown in the tables, our method achieves state-of-the-art performance among existing methods, even including UDA paradigms which under the easier setting. Moreover, on large-scale datasets like VisDA-2017 and DomainNet, our results significantly surpass the second-best one and achieve the best class-wise accuracy in most classes, indicating a general improvement for SFDA task.

In \cref{tab2}, we evaluate the generalization ability of our adaptive model. We reproduce previous works with open-source code and test them on VisDA-2017 target domain testing data mentioned above, and evaluate the performance drop between the target training and testing data. Results demonstrate that our consistency regularization-based paradigm shows high accuracy both in training and testing data, avoiding a giant performance degradation on testing data like what appeared in previous works. It proves that we successfully construct a smooth manifold and avoid overfitting issues, which is essential for model generalization and real-world SFDA applications. Note that UDA methods compared in \cref{tab2} are from VisDA2017 Classification Challenge~\cite{peng2017visda} with no constraint for model backbone, so unfair comparisons may exist, like challenge winner team ``GF\_ColourLab\_UEA''~\cite{french2017self} (not included in the table) choosing ResNet-152 as a backbone.

For the Office-Home dataset, as shown in \cref{tab3}, we achieve comparable results to previous state-of-the-art works. Consistent with what we discussed above, the use of strong image augmentations produces a harder optimization process, which calls for more training data to converge and boost model performance. With only 60 images per class on average, our method doesn't provide such significant improvement as that on large-scale datasets.

\subsection{Analysis}

\noindent \textbf{Ablation study.} 
We conduct ablation studies of our proposed modules in \cref{tab4}. Three class-wise accuracy shown here is carefully selected: class ``car'' owns the largest amount of samples while class ``knife'' and ``truck'' suffer from a more serious domain shift. Results in the first row are source-only model accuracy without adaptation. Row 2 and 3 show that a sampling-based pseudo-label selection strategy can significantly boost the consistency regularization learning paradigm because it identify noisy pseudo-labels and leverages potentially useful information. However, the accuracy of class ``truck'' decreases to 0\% due to the degradation issues mentioned above. Class balancing constraint by class-wise weighting module solves this problem properly, making sure that classes with severe domain shifts are correctly recognized. Lastly, the prototype calibration module can further enhance the adaptation effect by recognizing and filtering outliers.

\begin{table}[t]
	\centering
	\setlength{\abovecaptionskip}{2mm}
	\caption{Ablation study results of our proposed modules on VisDA-2017 dataset. Three typical class-wise and all classes' average accuracies are shown. (C.R.=Consistency Regularization, S.S.=Sampling-based pseudo-label Selection, C.W.=Class-wise Weighting, P.C.=Prototype Calibration)}
	%	\resizebox{2.1\columnwidth}{!}{
		\begin{tabular}{cccc|ccc|c}
			\toprule
			C.R. & S.S. & C.W. & P.C. & car & knife & truck & Avg. \\
			
			\midrule
			
			& & & & 75.7 & 4.4 & 7.9 & 46.6 \\
			$\checkmark$ & & & & 62.9 & 54.5 & 17.8 & 68.0 \\
			$\checkmark$ & $\checkmark$ & & & 89.4 & 97.6 & 0 & 84.9 \\
			$\checkmark$ & $\checkmark$ & & $\checkmark$ & 90.0 & 98.2 & 0 & 85.5 \\
			$\checkmark$ & $\checkmark$ & $\checkmark$ & & 65.3 & 98.9 & 60.7 & 88.0 \\
			$\checkmark$ & $\checkmark$ & $\checkmark$ & $\checkmark$ & 78.1 & 99.3 & 57.3 & \textbf{88.7} \\
			
			\bottomrule
		\end{tabular}
		%	}
	\label{tab4}
	\vspace{-2mm}
\end{table}

\noindent \textbf{Consistency regularization for boosting generalization ability.} 
\cref{fig:cr} demonstrates the effect of our proposed consistency regularization on boosting model generalization ability. As shown in \cref{fig:cr-1}, removing the consistency regularization module and applying vanilla self-training, which has no strong data augmentation, will remarkably reduce the model's final accuracy, especially on the target testing set. This is because the self-training approach fails to construct a smooth manifold on target domain data representations. According to \cref{fig:cr-2}, soft pseudo-labels are also essential to model generalization ability, because hard one-hot pseudo-labels may enlarge noise signals and cause error accumulation since there's no external supervision signal.

\noindent \textbf{Sampling-based pseudo-label selection.} 
We implement a vanilla thresholding method to select pseudo-label with various thresholds and compare them with our non-parametric sampling-based approach in \cref{fig:sampling}. While it's hard for thresholding to achieve high accuracy on all datasets at the same time, our proposed sampling-based method is free from adjusting any hyper-parameter and performs better for its consideration of more potentially useful samples. As shown in \cref{fig:sampling-2}, because limited samples are provided on the Office-Home dataset, our expectation approximation strategy further boosts the sampling-based selection effect. Note that on the VisDA-2017 dataset, the vanilla sampling way achieves the same results with expectation approximation due to efficient training samples.

\begin{table}[t]
	\centering
	\setlength{\abovecaptionskip}{2mm}
	\caption{Comparisons between our class-wise weighting method and vanilla class balancing loss, where $\beta$ is the loss weight. Best results inside each method are in bold font. Our approach achieves higher accuracy and shows less sensitivity to hyper-parameter.}
	\resizebox{0.8\columnwidth}{!}{
		\begin{tabular}{c|c|cc}
			\toprule
			Method & VisDA & Ar$\rightarrow$Cl & Ar$\rightarrow$Rw \\
			
			\midrule
			
			$\mathcal{L}_{div}$ ($\beta=1$) & 78.6 & 56.4 & 80.0 \\
			$\mathcal{L}_{div}$ ($\beta=0.5$) & 82.9 & \textbf{57.7} & \textbf{81.8} \\
			$\mathcal{L}_{div}$ ($\beta=0.1$) & 87.8 & 54.8 & 80.2 \\
			$\mathcal{L}_{div}$ ($\beta=0.05$) & \textbf{88.1} & 54.3 & 79.5 \\
			$\mathcal{L}_{div}$ ($\beta=0.01$) & 88.0 & 54.1 & 78.5 \\
			
			\midrule
			\midrule
			
			Ours ($\tau=0.9$) & 88.6 & 57.7 & 82.8 \\
			Ours ($\tau=0.8$) & \textbf{88.7} & \textbf{58.6} & 82.9 \\
			Ours ($\tau=0.6$) & 88.6 & 58.3 & \textbf{83.0} \\
			Ours ($\tau=0.3$) & 88.6 & 57.7 & 82.8 \\
			Ours ($\tau=0.1$) & 88.6 & 57.3 & 82.8 \\
			
			\bottomrule
		\end{tabular}
		}
	\label{tab5}
	\vspace{-3mm}
\end{table}

\noindent \textbf{Class-wise weighting for class balancing.} 
As what we discussed about \cref{tab4}, class-wise weighting can effectively balance class distribution and avoid trivial solution problems, and we further investigate the difference to widely used class balancing loss~\cite{liang2020we,yang2021exploiting,yang2021generalized} $\mathcal{L}_{div}$ (see \cref{Global-oriented-calibration}), shown in \cref{tab5}. Different from the high sensitivity to loss weight $\beta$ settings of $\mathcal{L}_{div}$, our method is robust to different hyper-parameter settings and outperforms $\mathcal{L}_{div}$-based methods without introducing any additional loss. Meanwhile, our method could achieve the best results while $\tau$ keeps consistent across tasks, where the $\mathcal{L}_{div}$ approach needs to fine-tune their hyper-parameters in a large range.

\noindent \textbf{Prototype calibration denoise effect.} 
\cref{fig:proto} demonstrates the effectiveness of prototypes in pseudo-label denoising. \cref{fig:proto-1} shows that with prototype calibration, pseudo-labels could achieve relatively high accuracy at the beginning stage of adaptation. As what we discussed in \cref{Global-oriented-calibration}, prototype calibration won't abandon potentially useful information due to its real-time property, and \cref{fig:proto-2} shows how the ratio of selected pseudo-labels varies along the training stage. It illustrates that during the adaptation process, an increasing number of pseudo-labels are selected, which means sample predictions are getting more consistent with cluster relation results in the feature space.

\begin{figure}[t]
	\centering
	\setlength{\abovecaptionskip}{2mm}
	\begin{subfigure}[b]{0.49\linewidth}
		\centering
		\includegraphics[width=\linewidth]{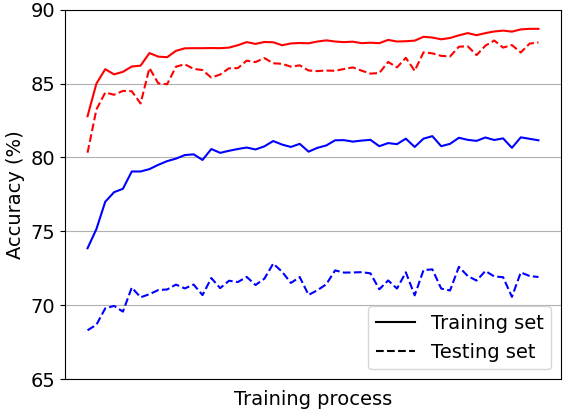}
		\caption{}
		\label{fig:cr-1}
	\end{subfigure}
	\hfill
	\begin{subfigure}[b]{0.49\linewidth}
		\centering
		\includegraphics[width=\linewidth]{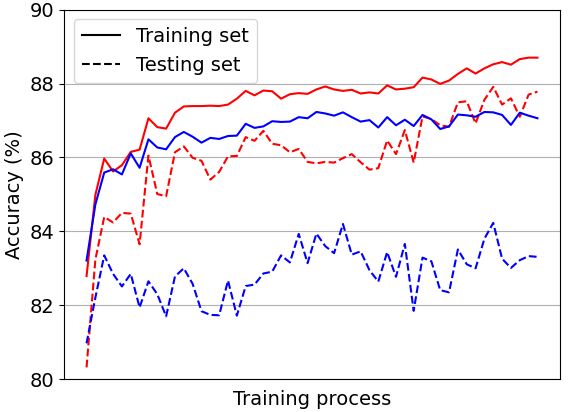}
		\caption{}
		\label{fig:cr-2}
	\end{subfigure}
        \vspace{-3mm}
	\caption{Accuracy on target training (real line) and testing set (dashed line). Results of our proposed consistency regularization (CR) with soft pseudo-label are in {\color{red} Red}, while {\color{blue} blue} lines indicate: (a) applying vanilla self-training instead of CR; (b) realizing CR with hard one-hot pseudo-labels.}
	\label{fig:cr}
	\vspace{-2mm}
\end{figure}

\begin{figure}[t]
	\centering
	\setlength{\abovecaptionskip}{2mm}
	\begin{subfigure}[b]{0.49\linewidth}
		\centering
		\includegraphics[width=\linewidth]{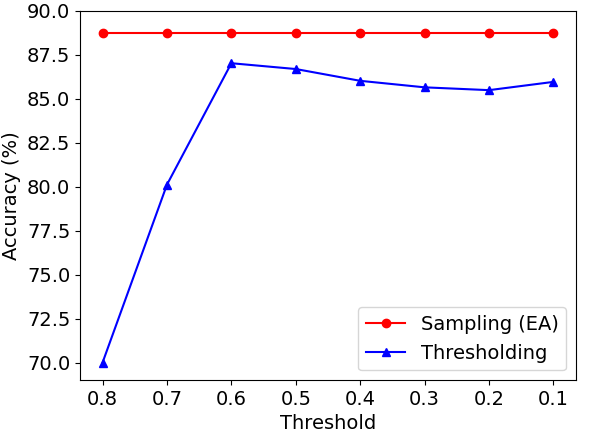}
		\caption{}
		\label{fig:sampling-1}
	\end{subfigure}
	\hfill
	\begin{subfigure}[b]{0.49\linewidth}
		\centering
		\includegraphics[width=\linewidth]{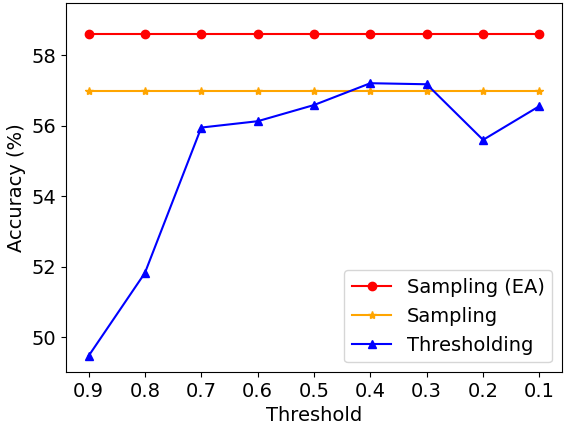}
		\caption{}
		\label{fig:sampling-2}
	\end{subfigure}
       \vspace{-3mm}
	\caption{Comparisons between our sampling-based pseudo-label selection and thresholding way with different thresholds on (a) VisDA-2017 (b) Ar$\rightarrow$Cl task of Office-Home. EA denotes the expectation approximation in \cref{eq4}.}
	\label{fig:sampling}
	\vspace{-2mm}
\end{figure}

\begin{figure}[t]
	\centering
	\setlength{\abovecaptionskip}{2mm}
	\begin{subfigure}[b]{0.49\linewidth}
		\centering
		\includegraphics[width=\linewidth]{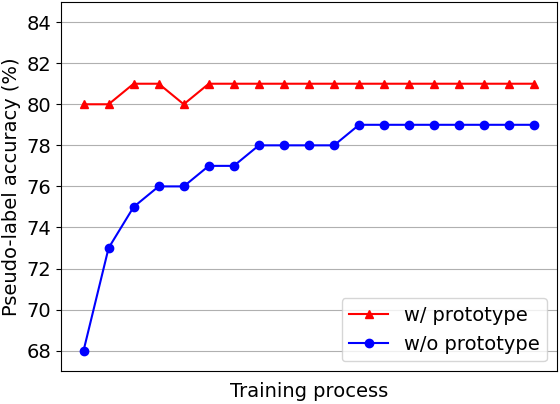}
		\caption{}
		\label{fig:proto-1}
	\end{subfigure}
	\hfill
	\begin{subfigure}[b]{0.49\linewidth}
		\centering
		\includegraphics[width=\linewidth]{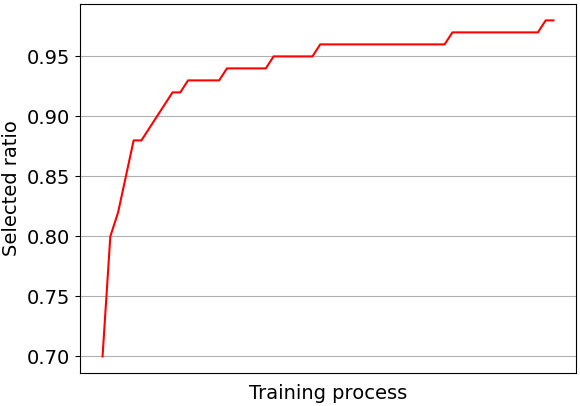}
		\caption{}
		\label{fig:proto-2}
	\end{subfigure}
       \vspace{-3mm}
	\caption{(a) Pseudo-label accuracy with and without prototype calibration on Ar$\rightarrow$Pr. (b) Ratio of selected pseudo-labels along the adaptation process on Ar$\rightarrow$Cl.}
	\label{fig:proto}
	\vspace{-2mm}
\end{figure}

\section{Conclusions}
\label{sec:conclu}

In this paper, based on the observation that previous SFDA works suffer from handling unseen but identically distributed data, we introduce a consistency regularization framework to achieve SFDA and boost model's generalization ability for the first time. By leveraging data augmentation and soft pseudo-label, our method can significantly improve the model performance both on target training and testing sets. In addition, sampling-based pseudo-label selection strategy and global-oriented calibration modules are proposed to further enhance the adaptation effect.
Experiments demonstrate the effectiveness of our method.

\vspace{1mm}

\noindent \textbf{Acknowledgment.}
This work was supported by Shenzhen Key Laboratory of next generation interactive media innovative technology (No.ZDSYS20210623092001004).

{\small
\bibliographystyle{ieee_fullname}
\bibliography{mybib}

\begin{thebibliography}{10}\itemsep=-1pt

\bibitem{berthelot2019mixmatch}
David Berthelot, Nicholas Carlini, Ian Goodfellow, Nicolas Papernot, Avital
  Oliver, and Colin~A Raffel.
\newblock Mixmatch: A holistic approach to semi-supervised learning.
\newblock {\em Advances in neural information processing systems}, 32, 2019.

\bibitem{caron2018deep}
Mathilde Caron, Piotr Bojanowski, Armand Joulin, and Matthijs Douze.
\newblock Deep clustering for unsupervised learning of visual features.
\newblock In {\em Proceedings of the European conference on computer vision
  (ECCV)}, pages 132--149, 2018.

\bibitem{cayton2005algorithms}
Lawrence Cayton.
\newblock Algorithms for manifold learning.
\newblock {\em Univ. of California at San Diego Tech. Rep}, 12(1-17):1, 2005.

\bibitem{cubuk2020randaugment}
Ekin~D Cubuk, Barret Zoph, Jonathon Shlens, and Quoc~V Le.
\newblock Randaugment: Practical automated data augmentation with a reduced
  search space.
\newblock In {\em Proceedings of the IEEE/CVF conference on computer vision and
  pattern recognition workshops}, pages 702--703, 2020.

\bibitem{cui2020gradually}
Shuhao Cui, Shuhui Wang, Junbao Zhuo, Chi Su, Qingming Huang, and Qi Tian.
\newblock Gradually vanishing bridge for adversarial domain adaptation.
\newblock In {\em Proceedings of the IEEE/CVF conference on computer vision and
  pattern recognition}, pages 12455--12464, 2020.

\bibitem{deng2021unbiased}
Jinhong Deng, Wen Li, Yuhua Chen, and Lixin Duan.
\newblock Unbiased mean teacher for cross-domain object detection.
\newblock In {\em Proceedings of the IEEE/CVF Conference on Computer Vision and
  Pattern Recognition}, pages 4091--4101, 2021.

\bibitem{devlin2018bert}
Jacob Devlin, Ming-Wei Chang, Kenton Lee, and Kristina Toutanova.
\newblock Bert: Pre-training of deep bidirectional transformers for language
  understanding.
\newblock {\em arXiv preprint arXiv:1810.04805}, 2018.

\bibitem{ding2022source}
Ning Ding, Yixing Xu, Yehui Tang, Chao Xu, Yunhe Wang, and Dacheng Tao.
\newblock Source-free domain adaptation via distribution estimation.
\newblock In {\em Proceedings of the IEEE/CVF Conference on Computer Vision and
  Pattern Recognition}, pages 7212--7222, 2022.

\bibitem{dosovitskiy2020image}
Alexey Dosovitskiy, Lucas Beyer, Alexander Kolesnikov, Dirk Weissenborn,
  Xiaohua Zhai, Thomas Unterthiner, Mostafa Dehghani, Matthias Minderer, Georg
  Heigold, Sylvain Gelly, et~al.
\newblock An image is worth 16x16 words: Transformers for image recognition at
  scale.
\newblock {\em arXiv preprint arXiv:2010.11929}, 2020.

\bibitem{eastwood2021source}
Cian Eastwood, Ian Mason, Chris Williams, and Bernhard Sch{\"o}lkopf.
\newblock Source-free adaptation to measurement shift via bottom-up feature
  restoration.
\newblock In {\em International Conference on Learning Representations}, 2021.

\bibitem{french2017self}
Geoffrey French, Michal Mackiewicz, and Mark Fisher.
\newblock Self-ensembling for visual domain adaptation.
\newblock {\em arXiv preprint arXiv:1706.05208}, 2017.

\bibitem{ganin2016domain}
Yaroslav Ganin, Evgeniya Ustinova, Hana Ajakan, Pascal Germain, Hugo
  Larochelle, Fran{\c{c}}ois Laviolette, Mario Marchand, and Victor Lempitsky.
\newblock Domain-adversarial training of neural networks.
\newblock {\em The journal of machine learning research}, 17(1):2096--2030,
  2016.

\bibitem{gretton2006kernel}
Arthur Gretton, Karsten Borgwardt, Malte Rasch, Bernhard Sch{\"o}lkopf, and
  Alex Smola.
\newblock A kernel method for the two-sample-problem.
\newblock {\em Advances in neural information processing systems}, 19, 2006.

\bibitem{gu2020spherical}
Xiang Gu, Jian Sun, and Zongben Xu.
\newblock Spherical space domain adaptation with robust pseudo-label loss.
\newblock In {\em Proceedings of the IEEE/CVF Conference on Computer Vision and
  Pattern Recognition}, pages 9101--9110, 2020.

\bibitem{he2023HQG}
Chunming He, Kai Li, Guoxia Xu, Jiangpeng Yan, Longxiang Tang, Yulun Zhang, Xiu
  Li, and Yaowei Wang.
\newblock Hqg-net: Unpaired medical image enhancement with high-quality
  guidance.
\newblock {\em arXiv preprint arXiv:2307.07829}, 2023.

\bibitem{he2023camouflaged}
Chunming He, Kai Li, Yachao Zhang, Longxiang Tang, Yulun Zhang, Zhenhua Guo,
  and Xiu Li.
\newblock Camouflaged object detection with feature decomposition and edge
  reconstruction.
\newblock In {\em Proceedings of the IEEE/CVF Conference on Computer Vision and
  Pattern Recognition}, pages 22046--22055, 2023.

\bibitem{he2023weakly}
Chunming He, Kai Li, Yachao Zhang, Guoxia Xu, Longxiang Tang, Yulun Zhang,
  Zhenhua Guo, and Xiu Li.
\newblock Weakly-supervised concealed object segmentation with sam-based pseudo
  labeling and multi-scale feature grouping.
\newblock {\em arXiv preprint arXiv:2305.11003}, 2023.

\bibitem{he2016deep}
Kaiming He, Xiangyu Zhang, Shaoqing Ren, and Jian Sun.
\newblock Deep residual learning for image recognition.
\newblock In {\em Proceedings of the IEEE conference on computer vision and
  pattern recognition}, pages 770--778, 2016.

\bibitem{huang2021rda}
Jiaxing Huang, Dayan Guan, Aoran Xiao, and Shijian Lu.
\newblock Rda: Robust domain adaptation via fourier adversarial attacking.
\newblock In {\em Proceedings of the IEEE/CVF International Conference on
  Computer Vision}, pages 8988--8999, 2021.

\bibitem{jin2020minimum}
Ying Jin, Ximei Wang, Mingsheng Long, and Jianmin Wang.
\newblock Minimum class confusion for versatile domain adaptation.
\newblock In {\em European Conference on Computer Vision}, pages 464--480.
  Springer, 2020.

\bibitem{laine2016temporal}
Samuli Laine and Timo Aila.
\newblock Temporal ensembling for semi-supervised learning.
\newblock {\em arXiv preprint arXiv:1610.02242}, 2016.

\bibitem{lee2013pseudo}
Dong-Hyun Lee et~al.
\newblock Pseudo-label: The simple and efficient semi-supervised learning
  method for deep neural networks.
\newblock In {\em Workshop on challenges in representation learning, ICML},
  volume~3, page 896, 2013.

\bibitem{li2021ecacl}
Kai Li, Chang Liu, Handong Zhao, Yulun Zhang, and Yun Fu.
\newblock Ecacl: A holistic framework for semi-supervised domain adaptation.
\newblock In {\em Proceedings of the IEEE/CVF International Conference on
  Computer Vision}, pages 8578--8587, 2021.

\bibitem{li2023source}
Kai Li, Deep Patel, Erik Kruus, and Martin~Renqiang Min.
\newblock Source-free video domain adaptation with spatial-temporal-historical
  consistency learning.
\newblock In {\em Proceedings of the IEEE/CVF Conference on Computer Vision and
  Pattern Recognition}, pages 14643--14652, 2023.

\bibitem{li2020adversarial}
Kai Li, Yulun Zhang, Kunpeng Li, and Yun Fu.
\newblock Adversarial feature hallucination networks for few-shot learning.
\newblock In {\em Proceedings of the IEEE/CVF conference on computer vision and
  pattern recognition}, pages 13470--13479, 2020.

\bibitem{li2021transferable}
Shuang Li, Mixue Xie, Kaixiong Gong, Chi~Harold Liu, Yulin Wang, and Wei Li.
\newblock Transferable semantic augmentation for domain adaptation.
\newblock In {\em Proceedings of the IEEE/CVF Conference on Computer Vision and
  Pattern Recognition}, pages 11516--11525, 2021.

\bibitem{liang2020we}
Jian Liang, Dapeng Hu, and Jiashi Feng.
\newblock Do we really need to access the source data? source hypothesis
  transfer for unsupervised domain adaptation.
\newblock In {\em International Conference on Machine Learning}, pages
  6028--6039. PMLR, 2020.

\bibitem{long2015learning}
Mingsheng Long, Yue Cao, Jianmin Wang, and Michael Jordan.
\newblock Learning transferable features with deep adaptation networks.
\newblock In {\em International conference on machine learning}, pages 97--105.
  PMLR, 2015.

\bibitem{long2018conditional}
Mingsheng Long, Zhangjie Cao, Jianmin Wang, and Michael~I Jordan.
\newblock Conditional adversarial domain adaptation.
\newblock {\em NeurIPS}, 2018.

\bibitem{lu2020stochastic}
Zhihe Lu, Yongxin Yang, Xiatian Zhu, Cong Liu, Yi-Zhe Song, and Tao Xiang.
\newblock Stochastic classifiers for unsupervised domain adaptation.
\newblock In {\em Proceedings of the IEEE/CVF Conference on Computer Vision and
  Pattern Recognition}, pages 9111--9120, 2020.

\bibitem{miyato2018virtual}
Takeru Miyato, Shin-ichi Maeda, Masanori Koyama, and Shin Ishii.
\newblock Virtual adversarial training: a regularization method for supervised
  and semi-supervised learning.
\newblock {\em IEEE transactions on pattern analysis and machine intelligence},
  41(8):1979--1993, 2018.

\bibitem{na2021fixbi}
Jaemin Na, Heechul Jung, Hyung~Jin Chang, and Wonjun Hwang.
\newblock Fixbi: Bridging domain spaces for unsupervised domain adaptation.
\newblock In {\em Proceedings of the IEEE/CVF Conference on Computer Vision and
  Pattern Recognition}, pages 1094--1103, 2021.

\bibitem{peng2019moment}
Xingchao Peng, Qinxun Bai, Xide Xia, Zijun Huang, Kate Saenko, and Bo Wang.
\newblock Moment matching for multi-source domain adaptation.
\newblock In {\em Proceedings of the IEEE/CVF international conference on
  computer vision}, pages 1406--1415, 2019.

\bibitem{peng2017visda}
Xingchao Peng, Ben Usman, Neela Kaushik, Judy Hoffman, Dequan Wang, and Kate
  Saenko.
\newblock Visda: The visual domain adaptation challenge.
\newblock {\em arXiv preprint arXiv:1710.06924}, 2017.

\bibitem{rozantsev2018beyond}
Artem Rozantsev, Mathieu Salzmann, and Pascal Fua.
\newblock Beyond sharing weights for deep domain adaptation.
\newblock {\em IEEE transactions on pattern analysis and machine intelligence},
  41(4):801--814, 2018.

\bibitem{saenko2010adapting}
Kate Saenko, Brian Kulis, Mario Fritz, and Trevor Darrell.
\newblock Adapting visual category models to new domains.
\newblock In {\em European conference on computer vision}, pages 213--226.
  Springer, 2010.

\bibitem{saito2019semi}
Kuniaki Saito, Donghyun Kim, Stan Sclaroff, Trevor Darrell, and Kate Saenko.
\newblock Semi-supervised domain adaptation via minimax entropy.
\newblock In {\em Proceedings of the IEEE/CVF international conference on
  computer vision}, pages 8050--8058, 2019.

\bibitem{salimans2016weight}
Tim Salimans and Durk~P Kingma.
\newblock Weight normalization: A simple reparameterization to accelerate
  training of deep neural networks.
\newblock {\em Advances in neural information processing systems}, 29, 2016.

\bibitem{sohn2020fixmatch}
Kihyuk Sohn, David Berthelot, Nicholas Carlini, Zizhao Zhang, Han Zhang,
  Colin~A Raffel, Ekin~Dogus Cubuk, Alexey Kurakin, and Chun-Liang Li.
\newblock Fixmatch: Simplifying semi-supervised learning with consistency and
  confidence.
\newblock {\em Advances in neural information processing systems}, 33:596--608,
  2020.

\bibitem{tang2020unsupervised}
Hui Tang, Ke Chen, and Kui Jia.
\newblock Unsupervised domain adaptation via structurally regularized deep
  clustering.
\newblock In {\em Proceedings of the IEEE/CVF conference on computer vision and
  pattern recognition}, pages 8725--8735, 2020.

\bibitem{tang2023source}
Longxiang Tang, Kai Li, Chunming He, Yulun Zhang, and Xiu Li.
\newblock Source-free domain adaptive fundus image segmentation with
  class-balanced mean teacher.
\newblock {\em arXiv preprint arXiv:2307.09973}, 2023.

\bibitem{tarvainen2017mean}
Antti Tarvainen and Harri Valpola.
\newblock Mean teachers are better role models: Weight-averaged consistency
  targets improve semi-supervised deep learning results.
\newblock {\em Advances in neural information processing systems}, 30, 2017.

\bibitem{venkateswara2017deep}
Hemanth Venkateswara, Jose Eusebio, Shayok Chakraborty, and Sethuraman
  Panchanathan.
\newblock Deep hashing network for unsupervised domain adaptation.
\newblock In {\em Proceedings of the IEEE conference on computer vision and
  pattern recognition}, pages 5018--5027, 2017.

\bibitem{wang2022exploring}
Fan Wang, Zhongyi Han, Yongshun Gong, and Yilong Yin.
\newblock Exploring domain-invariant parameters for source free domain
  adaptation.
\newblock In {\em Proceedings of the IEEE/CVF Conference on Computer Vision and
  Pattern Recognition}, pages 7151--7160, 2022.

\bibitem{wang2018deep}
Mei Wang and Weihong Deng.
\newblock Deep visual domain adaptation: A survey.
\newblock {\em Neurocomputing}, 312:135--153, 2018.

\bibitem{xia2021adaptive}
Haifeng Xia, Handong Zhao, and Zhengming Ding.
\newblock Adaptive adversarial network for source-free domain adaptation.
\newblock In {\em Proceedings of the IEEE/CVF International Conference on
  Computer Vision}, pages 9010--9019, 2021.

\bibitem{xie2020unsupervised}
Qizhe Xie, Zihang Dai, Eduard Hovy, Thang Luong, and Quoc Le.
\newblock Unsupervised data augmentation for consistency training.
\newblock {\em Advances in Neural Information Processing Systems},
  33:6256--6268, 2020.

\bibitem{xie2020self}
Qizhe Xie, Minh-Thang Luong, Eduard Hovy, and Quoc~V Le.
\newblock Self-training with noisy student improves imagenet classification.
\newblock In {\em Proceedings of the IEEE/CVF conference on computer vision and
  pattern recognition}, pages 10687--10698, 2020.

\bibitem{xu2020reliable}
Renjun Xu, Pelen Liu, Liyan Wang, Chao Chen, and Jindong Wang.
\newblock Reliable weighted optimal transport for unsupervised domain
  adaptation.
\newblock In {\em Proceedings of the IEEE/CVF conference on computer vision and
  pattern recognition}, pages 4394--4403, 2020.

\bibitem{xu2021cdtrans}
Tongkun Xu, Weihua Chen, Pichao Wang, Fan Wang, Hao Li, and Rong Jin.
\newblock Cdtrans: Cross-domain transformer for unsupervised domain adaptation.
\newblock {\em arXiv preprint arXiv:2109.06165}, 2021.

\bibitem{yang2021exploiting}
Shiqi Yang, Joost van~de Weijer, Luis Herranz, Shangling Jui, et~al.
\newblock Exploiting the intrinsic neighborhood structure for source-free
  domain adaptation.
\newblock {\em Advances in Neural Information Processing Systems},
  34:29393--29405, 2021.

\bibitem{yang2020casting}
Shiqi Yang, Yaxing Wang, Joost van~de Weijer, Luis Herranz, and Shangling Jui.
\newblock Casting a bait for offline and online source-free domain adaptation.
\newblock {\em arXiv preprint arXiv:2010.12427}, 2020.

\bibitem{yang2021generalized}
Shiqi Yang, Yaxing Wang, Joost van~de Weijer, Luis Herranz, and Shangling Jui.
\newblock Generalized source-free domain adaptation.
\newblock In {\em Proceedings of the IEEE/CVF International Conference on
  Computer Vision}, pages 8978--8987, 2021.

\bibitem{yang2021survey}
Xiangli Yang, Zixing Song, Irwin King, and Zenglin Xu.
\newblock A survey on deep semi-supervised learning.
\newblock {\em arXiv preprint arXiv:2103.00550}, 2021.

\bibitem{zhai2019s4l}
Xiaohua Zhai, Avital Oliver, Alexander Kolesnikov, and Lucas Beyer.
\newblock S4l: Self-supervised semi-supervised learning.
\newblock In {\em Proceedings of the IEEE/CVF International Conference on
  Computer Vision}, pages 1476--1485, 2019.

\bibitem{zhang2021flexmatch}
Bowen Zhang, Yidong Wang, Wenxin Hou, Hao Wu, Jindong Wang, Manabu Okumura, and
  Takahiro Shinozaki.
\newblock Flexmatch: Boosting semi-supervised learning with curriculum pseudo
  labeling.
\newblock {\em Advances in Neural Information Processing Systems},
  34:18408--18419, 2021.

\bibitem{zhang2021prototypical}
Pan Zhang, Bo Zhang, Ting Zhang, Dong Chen, Yong Wang, and Fang Wen.
\newblock Prototypical pseudo label denoising and target structure learning for
  domain adaptive semantic segmentation.
\newblock In {\em Proceedings of the IEEE/CVF conference on computer vision and
  pattern recognition}, pages 12414--12424, 2021.

\bibitem{zhang2022dac}
Ziyi Zhang, Weikai Chen, Hui Cheng, Zhen Li, Siyuan Li, Liang Lin, and Guanbin
  Li.
\newblock Divide and contrast: Source-free domain adaptation via adaptive
  contrastive learning.
\newblock In {\em Conference on Neural Information Processing Systems
  (NeurIPS)}, 2022.

\bibitem{zou2018unsupervised}
Yang Zou, Zhiding Yu, BVK Kumar, and Jinsong Wang.
\newblock Unsupervised domain adaptation for semantic segmentation via
  class-balanced self-training.
\newblock In {\em Proceedings of the European conference on computer vision
  (ECCV)}, pages 289--305, 2018.

\end{thebibliography}
}

\end{document}